\pdfoutput=1

\documentclass[11pt,anonymous,review]{article}

\usepackage[review]{acl}

\usepackage{times}
\usepackage{latexsym}

\usepackage[T1]{fontenc}

\usepackage[utf8]{inputenc}

\usepackage{microtype}

\usepackage{inconsolata}

\usepackage{times}
\usepackage{latexsym}
\usepackage{amsthm,amsmath,amssymb}
\usepackage{mathrsfs}
\usepackage{bm}
\usepackage{booktabs}
\usepackage{color}
\usepackage{graphicx}
\usepackage{multirow}
\usepackage{makecell}

\usepackage{float}
\usepackage{url}
\usepackage{xcolor, soul}
\definecolor{amethyst}{rgb}{0.6, 0.4, 0.8}
\sethlcolor{amethyst}

\usepackage{adjustbox}
\usepackage{fancyvrb}
\usepackage{caption}
\usepackage{pdfpages}
\usepackage{pifont}

\newcommand{\squishlist}{
 \begin{list}{$\bullet$}
  { \setlength{\itemsep}{0pt}
     \setlength{\parsep}{3pt}
     \setlength{\topsep}{3pt}
     \setlength{\partopsep}{0pt}
     \setlength{\leftmargin}{1.5em}
     \setlength{\labelwidth}{1em}
     \setlength{\labelsep}{0.5em} } }

\newcommand{\squishend}{
  \end{list}  }

\title{Combining Hierachical VAEs with LLMs for clinically meaningful timeline summarisation in social media}

\author{Jiayu Song$^*$$^{1}$, Jenny Chim$^*$$^{1}$, Adam Tsakalidis$^{1,3}$, Julia Ive$^{1}$, Dana Atzil-Slonim$^{2}$, Maria Liakata$^{1,3}$ \\
        $^1$ Queen Mary University of London, London, UK \\ 
        $^2$ Bar-Ilan University, Israel\\ 
        $^3$ The Alan Turing Institute, London, UK\\
        \texttt{\{jiayu.song,c.chim,a.tsakalidis,j.ive,m.liakata\}@qmul.ac.uk}\\
        {dana.slonim@gmail.com}\\}

\begin{document}
\maketitle

\def\thefootnote{*}\footnotetext{Equal contribution.}\def\thefootnote{\arabic{footnote}}

\begin{abstract}

We introduce a hybrid abstractive summarisation approach combining hierarchical VAE with LLMs (LlaMA-2) to produce clinically meaningful summaries from social media user timelines, appropriate for mental health monitoring. 
The summaries combine two different narrative points of view: clinical insights in third person useful for a clinician are generated by feeding into an LLM specialised clinical prompts, and importantly, a temporally sensitive abstractive summary of the user's timeline in first person, generated by a novel hierarchical variational autoencoder, TH-VAE.
We assess the generated summaries via automatic evaluation against expert  summaries and via human evaluation with clinical experts, showing that timeline summarisation by TH-VAE results in more factual and logically coherent summaries rich in clinical utility and superior to LLM-only approaches in capturing changes over time.

\end{abstract}

\section{Introduction}

Social media users discuss different aspects of their lives, providing important clues about their mental health.
Previous work \cite{de2013predicting, coppersmith2014quantifying, cohan2018-smhd, chancellor2020methods} have studied users' social media posts to help identify depression, bipolar disorder \cite{yates2017-depression,husseini2018-deep} or self-harm \cite{zirikly2019clpsych}, and there have been efforts on multi-task learning to capture user states at a particular moment in time \cite{benton-etal-2017-multitask, yang-etal-2023-towards}. Despite the importance of longitudinal assessments of linguistic and other digital content for mental health clinical outcomes \cite{velupillai2018using}, there is little work on considering the evolution of an individual's mental health over time through their social media. 
\citet{tsakalidis2022-identifying, tsakalidis2022-overview} established the task of capturing changes (switches and escalations) in an individual's mood over time and showed how identifying these helps predict clinical assessments of suicidal ideation. 
However, currently clinicians don't have access to such information to assess individuals' mental-state and they mainly rely on self-reports completed by patients throughout psychotherapy \cite{crits2021psychotherapy}. 
Although standardized subjective measures are fundamental to mental health monitoring and research, they have significant limitations, such as the extent of individuals’ self-awareness, their willingness to complete questionnaires, and the limited choice of responses \cite{kazdin2021extending}. 
Providing concise summaries that can capture fluctuations in individuals' state-of-mind while emphasizing key clinical concepts, can significantly assist in monitoring, prevention and early detection of mental health issues. Such summaries would augment clinician capacity, provide alternatives to standard questionnaires and compensate for reduced access to mental health services \cite{schwartz2023assessments}. 

To the best of our knowledge we are the first to propose clinically meaningful summaries of social media user `timelines' (sequences of chronologically ordered posts by a user). Driven by the need to concisely summarise time-series language data which can span arbitrary lengths that exceed limits of many contemporary models and render purely extractive methods impractical, we propose a novel hybrid unsupervised abstractive method, Timeline Hierarchical VAE (TH-VAE). Our system makes use of a hierarchical variational autoencoder that compresses timeline information into compact representations and a large language model (LLM), creating a two-layer summary that combines two different narrative points of view. Specifically: a \textit{high-level} summary in third person useful for a clinician, is generated by feeding into an LLM specialised clinical prompts and importantly a temporally sensitive abstractive summary of the user's timeline in first person (\textit{evidence summary}), generated by TH-VAE. 
The generation of the first person abstractive evidence summary via TH-VAE is guided by mental health related key-phrases obtained through instruction prompting by an LLM.
The final resulting high level summary covers aspects considered to be crucial by clinicians from a wide range of therapeutic approaches, including individuals' diagnosis, intrapersonal and interpersonal patterns and extent of mental state changes over time \cite{eells2022handbook}.\footnote{For a complete list, please see Table \ref{tab:appendix.clinical_concepts} in Appendix \ref{sec:appendix}.} 

\noindent We make the following contributions:

\squishlist
\item We develop a novel abstractive timeline summarisation method (TH-VAE) based on adapting a hierarchical VAE model (NVAE)(\S\ref{method:h_vae}) to longitudinal social media data (user timelines). 

\item We provide a new task, the creation of clinically meaningful summaries from social media data. These summaries, generated in a hybrid approach, comprise high-level information in third person consistent with clinical insights (diagnosis, inter- and intra- personal aspects, moments of change) and evidence from a user's timeline, generated from the TH-VAE, supporting the assigned high-level insights. (\S\ref{method}) 

\item We create a dataset of expert-written mental health summaries from longitudinal social media data. A small sample of these is used to help with modeling (\S\ref{method:summary}) and the rest is used for evaluation (\S\ref{evaluation}).

\item We provide a novel detailed evaluation method of the summaries based on preservation of clinical information, summary consistency, and usefulness to clinicians, using semantic similarity based metrics, NLI based inference, as well as expert human evaluation 
(\S\ref{evaluation}).

\item We conduct experiments using different unsupervised summarisation methods based on LLMs and story generation (\S\ref{exp:baseline}), showing superior performance for TH-VAE (\S\ref{sec:results}).

\squishend

\section{Related Work}


\noindent\textbf{Timeline summarization}\label{timeline} aims at concisely summarizing the evolution trajectory of a 
specific topic along a timeline \cite{chen2019learning, chen2023follow} and has primarily focussed on news datasets. 
Methodologically it has involved both extractive and abstractive methods; for example, \citet{Allan2001new} define temporal summaries by extracting a sentence per event in a news story while \citet{li-2021-timeline} construct a multi-document event graph to capture long distance dependencies between events, weight events and extract an event summary sentence with maximum event coverage. 
\citet{li-timelinegeneration14, chang2016timeline, wang2021bringing, hills2023creation} detect important events in an individual's timeline and explore the event trajectory. 
In \citet{ren2013personalized} timeline summarisation involves identifying users' interests by defining a social circle from a set of friends and selecting salient tweets to obtain an extractive summary. \citet{chang2016timeline} also uses extractive summarisation and selects sentences based on different features (e.g., popularity-based, temporal).
Work in abstractive timeline summarisation \cite{martschat2018temporally,steen-19abstractive} involves identifying clusters of news or events to generate abstractive summaries from, or memory-based timeline summarisation to track the trajectory of events \cite{chen2019learning}.
By contrast we consider a user's timeline, a series of posts shared by an individual over a period of time \cite{tsakalidis2022-identifying}. Such timelines do not exhibit obvious or consistent topics, contain few events and an explosion of emotions. Our goal in user timeline summarisation is to capture important information and synthesise it.\\ 
\noindent\textbf{Summaries in Mental Health.}
Although summaries are clinically crucial for compiling information about individuals, there is limited literature on the subject, with the primary focus being on expert-generated case study summarization \cite{eells2022handbook}.
Only recently, researchers have started to use NLP capabilities to automatically generate summaries in the clinical domain. 
\citet{info:doi/10.2196/20865} demonstrated the usefulness of generating summarised diagnoses from a single-session interview. 
\citet{Srivastava20222Counseling} summarised psychotherapy conversations at the level of single counseling sessions proposing that summaries should exploit domain knowledge and conversational elements. 
On social media, \citet{sotudeh-etal-2022-mentsum} generated summaries of individual Reddit posts, relying on formatting conventions (i.e. TLDR) to extract short summaries provided by the users themselves without further content constraints. \citet{yang-etal-2023-towards} instruction-tuned LLMs to generate mental health analyses from static social media text. By contrast our work summarises user timelines and combines information from social media posts based on high-level expert domain knowledge, important for evaluating individuals' progression over time. \\
\noindent\textbf{Summarising with LLMs.} 
Current work on LLM-based summarisation focuses on news articles or instructional texts \cite{goyal2022zeroshotnews, https://doi.org/10.48550/arxiv.2301.13848, maynez-etal-2023-benchmarking}, using simple prompts (e.g. “Summarize the
following article:”). \citet{wang-etal-2023-element} took a multi-step approach, extracting event information from news via curated guiding questions then summarising the prompted outputs. In our work, we summarise longitudinal user generated content and use clinically-informed prompts to generate high-level mental health observations.\\
\noindent\textbf{Summary Evaluation.} 
Existing mental health summarisation works utilised natural language generation metrics, for example using ROUGE \cite{lin-2004-rouge} to measure n-gram overlap against reference documents \citep{info:doi/10.2196/20865, Srivastava20222Counseling, sotudeh-etal-2022-mentsum}. \citet{Srivastava20222Counseling} additionally applied BLEURT \cite{scialom-etal-2021-questeval}, a learned metric trained on ratings, QuestEval \cite{scialom-etal-2021-questeval}, a metric based on question generation and answering, and MHIC, a metric that they defined to assess information captured in counselling summaries based on ROUGE. 

\indent  Contrary to prior work, our task involves two-layer mental health summaries combining first-person social media content with high-level clinical concepts in third person, posing unique evaluation challenges. For example, data noisiness makes metrics learned on well-formed texts unsuitable, and evaluation must assess consistency both between summary layers and within the detailed high-level summary itself. To this end, we extend the line of work leveraging natural language inference (NLI) in summary factuality and consistency evaluation \cite{maynez-etal-2020-faithfulness, laban-etal-2022-summac}.
\section{Methodology}\label{method}

\textbf{Task}\label{task}
 Given a user's timeline (a series of posts between two dates \cite{tsakalidis2022-identifying}), the goal is to generate an abstractive summary that reflects the user's mental state and how it changes over time. This summary includes high-level information useful for clinicians in third person, and corresponding evidence from the timeline in first person.

\begin{figure}
\centering
\includegraphics[width=\columnwidth]{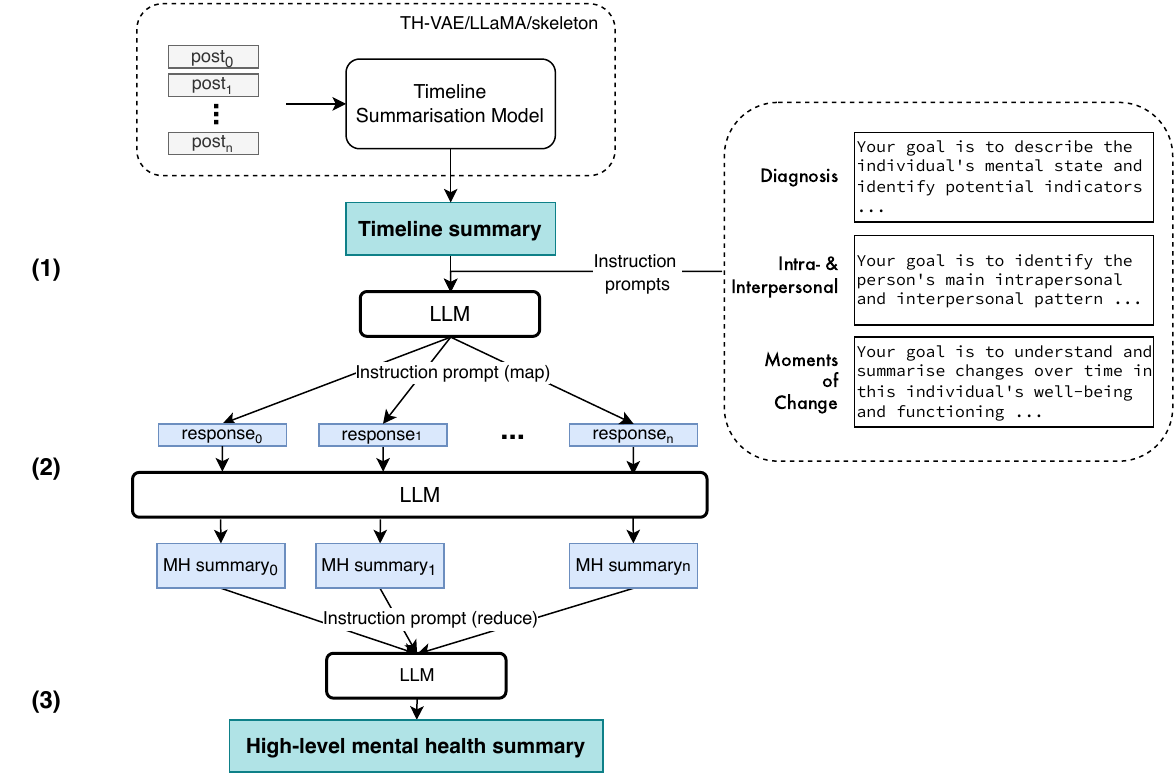}
\caption{Prompting framework for generating high-level summaries. Taking a first-person summarised timeline as input, we (1) prompt the LLM around different clinical topics, (2) summarise extracted inferences into prose per topic, and (3) combine the topic-specific intermediate summaries into a coherent, distilled document.} 
\label{method.fig:high_level_prompt}
\vspace{-1em}
\end{figure}

\subsection{Architecture Overview}\label{architecture_overview}
Fig.~\ref{method.fig:high_level_prompt} shows the summary generation process. It consists of two sub-processes:\\
(1) Abstractive generation of the timeline/evidence summary (\S\ref{method:h_vae}). We use three different unsupervised methods for creating the timeline summary in first person: Timeline hierarchical VAE (TH-VAE \S\ref{task:segments}), our key methodological novelty; LLaMA (\S\ref{exp:baseline}); a method from story generation (\S\ref{exp:baseline}).\\
(2) Generation of the High-level summary (\S\ref{method:summary}). We feed the generated timeline/evidence summary into an instruction-tuned LLM (Llama), where prompts originate from a small sample of expert human annotation (\S\ref{task:segments}), and generate high-level summaries covering clinical aspects such as diagnosis, inter- and intra- personal relationships and fluctuations in mood. 
The following subsections describe our novel timeline summarisation method using an adapted hierarchical VAE (TH-VAE).

\begin{figure}
\centering
\includegraphics[width=.8\linewidth]{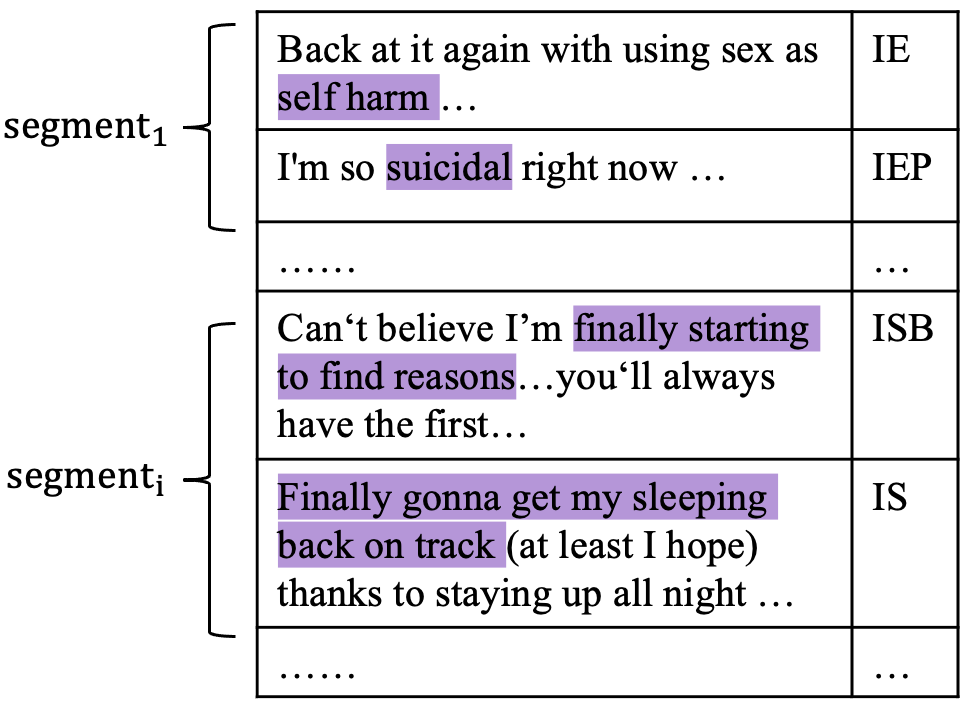}
\caption{Each timeline is separated into several segments based on 'MoC'. We \hl{highlight} the key phrases.}
\label{timeline}
\vspace{-0.5cm}
\end{figure}

\begin{figure*}
\centering
\includegraphics[width=.88\textwidth]{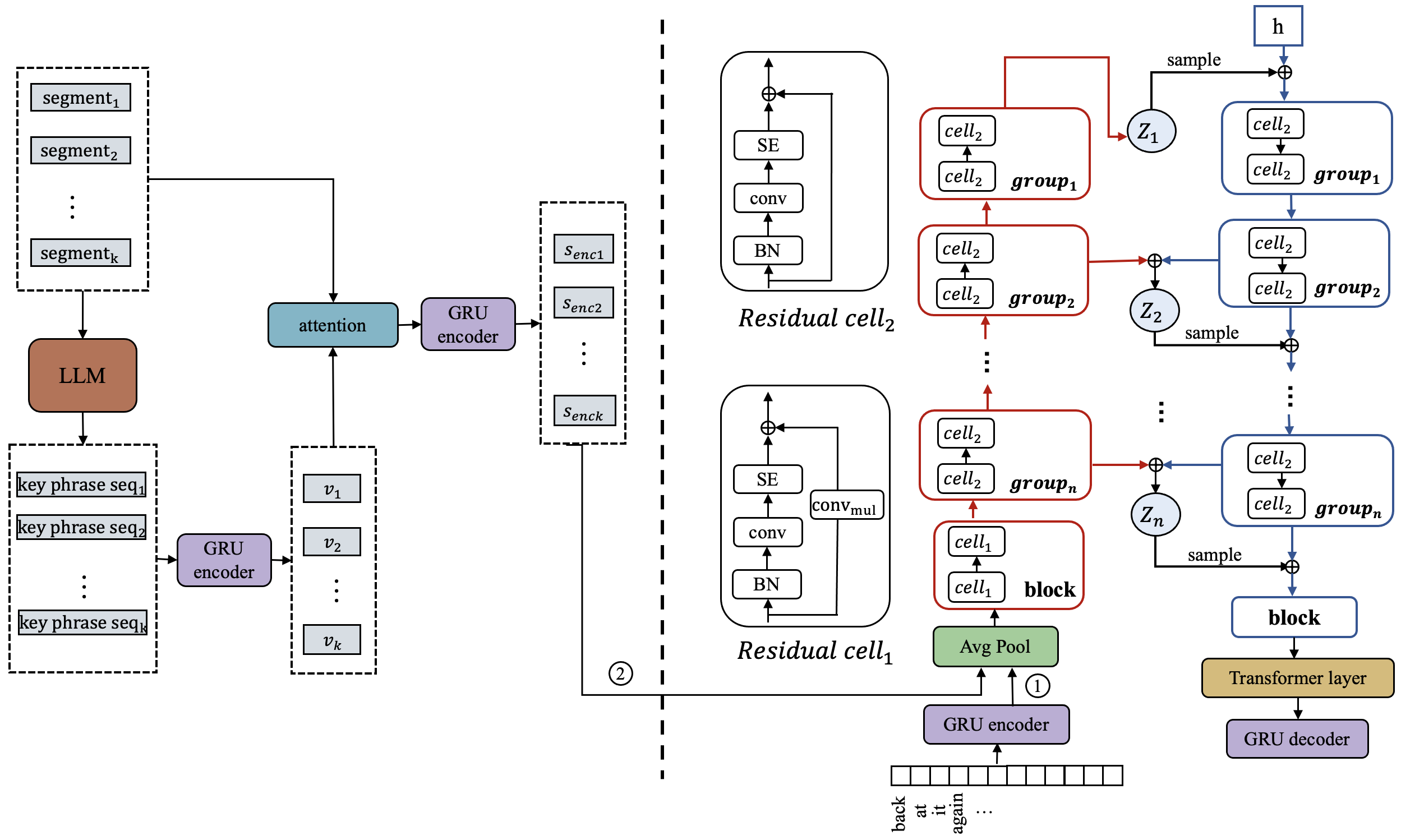}
\caption{Overview of TH-VAE. The left of the dotted line shows the construction of the k-sentence representation used only during generation, informed by the key-phrases, while the right side shows the hierarchical structure of TH-VAE, and its components.\ding{172} and \ding{173} represent the input during training and generation respectively.}
\vspace{-1em}
\label{fig:h-vae}
\end{figure*}

\subsection{Input to Timeline Summarisation}\label{task:segments}
The input to TH-VAE and the other timeline summarisation methods is a user's timeline, annotated with Moments of Change in mood (MoC)\cite{tsakalidis2022-identifying}. MoC annotations consist of \textit{Switches} (sudden mood shifts, denoted by `IS'--In Switch-- and `ISB'--In Switch Beginning-- tags),  and \textit{Escalations} (gradual mood progression, denoted by `IE'--In Escalation-- and `IEP'--In Escalation Peak-- tags).
We split the whole timeline (see Fig.~\ref{timeline}) into several segments (sub-timelines) based on 'MoC', 
so that consecutive posts with the same label (`IE' or `IEP'),(`ISB' or `IS') or `0' are grouped together. This assumes each segment consists of posts of a similar mood type, which facilitates capturing different features and relations between them. This is somewhat similar to news timeline summarisation which clusters around stories or events, with the additional challenge that mood features are more evasive and we hope to model these through latent variables.\\
\textbf{Key phrases}\label{method:key} We asked clinical psychologists to annotate key phrases indicative of users' mental health in three timelines. These phrases include mood related clues but also information on interpersonal relationships, behaviors or events related to a user's mental state (see highlights in Fig.~\ref{timeline}). We take these annotated timeline/key phrases pairs as examples and prompt LLaMA \cite{touvron2023llama} to annotate the rest of the timelines. \\
\textbf{Timeline summary representation}\label{method:summary_rep} For each segment ${s}_{i}$, we input its corresponding key phrase sequence $\left\{\boldsymbol{e}_{1}, ... , \boldsymbol{e}_{j}, ... , \boldsymbol{e}_{n}\right\}$ into a GRU encoder~\cite{cho2014learning} to get the key phrases encoding ${\boldsymbol{v}}=GRU(\left[\boldsymbol{e}_{1}; ... ;\boldsymbol{e}_{n}\right])$, which is represented by the last hidden state of the GRU (see left part of Fig\ref{fig:h-vae}). 
We calculate the similarity between ${\boldsymbol{v}}$ and each word embedding $\boldsymbol{w}_{i}$ in the segment as the weight $\alpha_{i}$: 
\vspace{-.5em}
\begin{equation*}
    \begin{aligned}
    {\alpha_{i}} = \frac{cos(\boldsymbol{v}, \boldsymbol{w}_{i})}{\sum_{i^{\prime}=1}^{m}cos(\boldsymbol{v}, \boldsymbol{w}_{i^{\prime}})}.
    \end{aligned} 
\end{equation*}

Thus ${s}_{i}$ can be represented by a series of weighted word embeddings $\left\{\alpha_{1}\boldsymbol{w}_{1}, \alpha_{2}\boldsymbol{w}_{2}, ... , \alpha_{m}\boldsymbol{w}_{m} \right\}$, where ${m}$ is the length of ${s}_{i}$. We encode it with the GRU encoder to get the segment representation ${\boldsymbol{s}_{enc_{i}}}$=$GRU(\left[\alpha_{1}\boldsymbol{w}_{1}; \alpha_{2}\boldsymbol{w}_{2}; ... ; \alpha_{m}\boldsymbol{w}_{m} \right])$. 
If the timeline is divided into $k$ segments, we can get $k$ segment encodings ${s}_{enc_1},...,{s}_{enc_k}$ in this way. We concatenate these encodings in chronological order to get a segment sequence $\left\{\boldsymbol{s}_{{enc}_{1}}, \boldsymbol{s}_{{enc}_{2}}, ... , \boldsymbol{s}_{{enc}_{k}}\right\}$, apply an average pooling operation (Avg Pool)\cite{Lin2013NetworkIN} over the output of the GRU encoder~(See right part of Fig\ref{fig:h-vae}) and feed it into the hierarchical part of TH-VAE to generate a timeline summary.  

\subsection{Overview of TH-VAE}
Due to the lack of gold summaries for training purposes, we have to construct the summary distribution without any guidance. 
Thus we need a model that can learn an expressive distribution for a long timeline (the longest timeline has 124 posts, and the longest of these posts has over 300 words). We also need to construct a mental health related summary distribution that can capture different features and establish the long-range dependencies between these features in the timeline.
We propose TH-VAE, an unsupervised abstractive timeline summarization model adapted from NVAE \cite{vahdat2020nvae}, to construct a more expressive prior for a user timeline. 

\vspace{-0.05cm}
In the learning process, we split the timeline into several segments (sub-timelines, \S\ref{method:h_vae}), considered to contain consecutive posts with similar mood, and train TH-VAE to learn the distribution of each segment ${s}$ by reconstructing it.


\vspace{-0.05cm}
 When generating the evidence summary, we still treat each segment as a unit. To help the model focus on important information during generation we introduce the notion of key phrases (\S\ref{method:key}). We use an automatic method based on an LLM to extract mental health related key phrases from each segment and encode key phrase-segment pairs with an attention mechanism. We concatenate the sequence of segment representations of a timeline in chronological order and input it into the hierarchical structure of TH-VAE to generate the timeline/evidence summary (See~ Fig.~\ref{fig:h-vae}, left part).

\vspace{-0.3cm}
\subsection{Document Reconstruction via TH-VAE}\label{method:h_vae}

The vanilla VAE assumes a prior $p(\boldsymbol{z})$ of document $\boldsymbol{x}$ over latent variables $\boldsymbol{z}$ to be a Normal Gaussian distribution, and parameterizes an approximate posterior distribution ${q_{\phi}(\boldsymbol{z}|\boldsymbol{x})}$ given text $\boldsymbol{x}$. It uses KL (Kullback–Leibler divergence) to calculate the distance between $p(\boldsymbol{z})$ and ${q_{\phi}(\boldsymbol{z}|\boldsymbol{x})}$ and gradually reduces the distance between them in training. Finally, it samples from the hypothesised posterior distribution and generates the document $\boldsymbol{x}$. It has been shown that the vanilla VAE can lead to over-regularising the posterior distribution, resulting in latent representations that do not represent well the structure of the data \cite{klushyn2019learning, alemi2018fixing, sonderby2016ladder, ranganath2016hierarchical, vahdat2020nvae}.
However, for a long document assuming its distribution to be a Gaussian does not provide enough expressive power; we need to be able to consider the structure of different semantic elements and the relationship between them.

The deep Hierarchical VAE (NVAE) \cite{vahdat2020nvae}, introduced for images, increases expressiveness by introducing several latent variables to generate large high-quality images, demonstrates the superiority of the hierarchical VAE.
Here, we adapt this model for long documents, resulting in Timeline Hierarchical VAE (TH-VAE), and use it as the basis of constructing mental health related timeline representations.

\subsubsection{Hierarchical Component}\label{sec:hierarchical}
TH-VAE increases the expressiveness of the approximate posterior and prior by partitioning the latent variable $\boldsymbol{z}$ into $l$ latent variables  $\boldsymbol{z}$=$\left\{\boldsymbol{z}_{1}, \boldsymbol{z}_{2}, ... , \boldsymbol{z}_{l} \right\}$\cite{vahdat2020nvae}. The prior is represented by ${{p(\boldsymbol{z})}={\prod_{l}{p(\boldsymbol{z}_{l}|\boldsymbol{z}_{<l})}}}$ and it parameterises the approximate posterior distribution ${q_{\phi}(\boldsymbol{z}|\boldsymbol{x})}={\prod_{l}{q_{\phi}(\boldsymbol{z}_{l}|\boldsymbol{z}_{<l},\boldsymbol{x})}}$ which are represented by factorial Normal distributions. This objective is to maximise its lower bound as:
\begin{equation*}
    \begin{aligned}
    {L(\theta;\boldsymbol{x})} = {-KL({q_{\phi}(\boldsymbol{z}_{1}|\boldsymbol{x})||p(\boldsymbol{z}_{1})})}\\
    \sum_{l=2}^{L}{\mathbb{E}_{q_{\phi}(\boldsymbol{z}<l|\boldsymbol{x})}\lbrack {-KL({q_{\phi}(\boldsymbol{z}_{l}|\boldsymbol{x},\boldsymbol{z}<l) || p(\boldsymbol{z}_{l}|\boldsymbol{z}<l)})} \rbrack}\\
    +{\mathbb{E}_{q_{\phi}(\boldsymbol{z}|\boldsymbol{x})}\lbrack {\log{p_{\theta}(\boldsymbol{x}|\boldsymbol{z})}} \rbrack}.\\ \label{vae_loss}
    \end{aligned}
    \vspace{-0.5cm}
\end{equation*}
\noindent Before going into the hierarchical architecture, we use a GRU encoder to encode the segment, to reduce the impact from padding. Then we add an Avg Pool \cite{Lin2013NetworkIN} over the output of the GRU encoder to fix the input length. Both TH-VAE and NVAE use multiple residual cells to construct the hierarchical structure. In TH-VAE we simplify residual cells to work with textual data rather than images, and keep the optimization strategies in NVAE, i.e., BN (batch normalization) with Swish Activation and Squeeze and Excitation (SE). We use two different residual cells: $residual\,cell_{1}$ and $residual\,cell_{2}$.  
The input representations first go through a \textbf{block} which focuses on capturing the features of a segment and consists of $residual\,cell_{1}$. 
To form $residual\,cell_{1}$ we use series BN, $conv$ (CNN with one kernel size), SE as well as $conv_{mul}$ (CNN with multiple kernel sizes), where the latter helps with capturing the different features. 
Then, the output of the \textbf{block} will go into the layered \textbf{group}s (see Fig.~\ref{fig:h-vae}--right), responsible for learning to capture the relationship between different features in segments and long-range dependencies between them. 
Each \textbf{group} is used to encode the sub-latent variables $\textbf{z}_{i}$ and consists of $residual\,cell_{2}$s. Since $conv_{mul}$ increased parameters without added benefit, we only use $conv$ in $residual\,cell_{2}$. Finally we add another \textbf{block} to integrate information.
During training, the whole hierarchical architecture is used to learn the distribution of each segment, by learning features and long-range dependencies within them via segment reconstruction (as shown in the right part of Fig\ref{fig:h-vae}). Then during generation a sequence of segments (a whole timeline) is input to TH-VAE to generate similarly structured text. The left part of Fig\ref{fig:h-vae} shows the process of encoding the sequence of segments. When decoding, we use the same decoder component as in \cite{song2022unsupervised}, comprising a transformer decoder (we load pre-trained parameters from BART) followed by a GRU decoder. 

\subsection{High-level Mental Health Summarization}\label{method:summary}
We focus on information considered important in summarising individuals' mental states according to therapeutic approaches \cite{eells2022handbook}. Although all users broadly talk about mental health related topics in this dataset, the extent to which clinical concepts appear in each one varies due to natural individual differences. As such, when annotators write gold summaries and when we generate model-written ones, we focus on  clinical information that is present, ignoring true negatives.

We prompt an instruction-tuned LLM following a multi-stage framework (Fig.~\ref{method.fig:high_level_prompt}) to generate high-level mental health summaries based on timeline summaries. In the \textit{map} stage, we instruct the model to provide inferences based on the timeline summary focusing on clinical topics (Appendix \ref{sec:appendix}, Table \ref{tab:appendix.clinical_concepts}), such as presenting issues, inter/intra-personal patterns, and moments of change. Instructions and prompts are in Appendix B. In the \textit{reduce} stage, we iteratively prompt the model to synthesise extracted observations into a concise summary. 
\vspace{-1em}
\section{Experiments}
\subsection{Evaluation Dataset Creation} \label{sec:experiments.dataset}
We work with three clinical psychology graduate students who are fluent in English to create gold evidence-supported summaries. We use the dataset collected by \citet{tsakalidis2022-identifying} comprising 500 anonymised user timelines from Talklife. The number of posts in each timeline varies ([12-124]). We sample 30 timelines for annotators to highlight information related to individuals' mental states and write high-level summaries which include diagnosis, intra- and interpersonal patterns and mental state changes over time. We use these for evaluation and 3 additional held out timelines for development and in-context learning key phrase extraction.

\subsection{Models \& Baselines}\label{exp:baseline}
We compare our method against existing models for unsupervised abstractive opinion summarisation. For experiment settings, model specifications, and prompts refer to Appendix~\ref{sec:appendix} and B.

\smallskip
\noindent\textbf{Skeleton-based model} is an unsupervised method proposed for story generation which encodes the skeleton (phrases that express the key meaning of sentences) to generate a detailed and polished sentence \cite{2018-skeleton}. We include it as one of the models to compare against as like TH-VAE it uses key phrases to generate a story/timeline in an unsupervised way. The key phrases provided are the same as for TH-VAE.

\smallskip
\noindent\textbf{LLaMA} 
We prompt a LLM to extract key phrases and then write TLDR-type summaries \cite{volske-etal-2017-tl} focusing on the key phrases. 
Resulting summaries are similar to concise user-authored ones commonly found in social media data. 


\smallskip
\noindent\textbf{High-level Summary} 
To obtain corresponding mental health summaries, we feed timeline summaries generated via TH-VAE and the above baselines into the LLM prompting framework outlined in \S\ref{method:summary}. In addition, to see the benefits of timeline summarisation and specific clinical prompts, we implement a high-level and prompt-only \textbf{naive baseline}. It involves splitting timelines into chunks, prompting the LLM to write a mental health summary of each chunk, and rewriting the chunk-level summaries into a single coherent document.
\begin{table*}[]
\centering
\begin{adjustbox}{width=\textwidth}
\begin{tabular}{ll}
\toprule
Timeline Summary & High-level Summary \\ \midrule
\begin{minipage}[t]{.9\columnwidth}%
I hate school. im so scared.i love the cat so much. i'm gonna go be counter for a while. unknown good night all! i hate my nose. fuck the sat. i got a commitment ring 3 i've got counterbodied under my eyes. i feel alone i feel like everyone hates me. i hate saying this because it sounds bad, but i really want someone to pay attention to me. solid my mom only cares about my brothers learning disabilities and her new boyfriend expressive. my boyfriend goes throug of ptsd like depression symptoms, so he gets distant sometimes levels. i literally have no friends argue. i've always been quiet but sometimes i just want someone never pay attention to me. i argue to know only when't be a body. i have to know. i'll have to be happy. %
\end{minipage}
& \begin{minipage}[t]{2\columnwidth}%
The individual is likely experiencing a range of mental health issues, including depression, anxiety, and low self-esteem. They report feeling isolated, lonely, and frustrated, with a history of stressful life events. The individual's mood appears to be low, with expressions of sadness, frustration, and anger. \\

It is evident that they have a profound craving for attention and connection with others, as well as a yearning for acceptance and validation. However, their fear of rejection and abandonment hinders them from forming and maintaining healthy relationships. Moreover, their self-criticism and preoccupation with perceived shortcomings indicate a lack of self-compassion and acceptance of their strengths and vulnerabilities.\\

The individual's mood and well-being have been observed to fluctuate over time, with both positive and negative changes experienced. The individual has expressed a range of emotions, including sadness, loneliness, and frustration, as well as moments of happiness and positivity. Noteworthy positive changes include their excitement about having a cat and receiving a commitment ring, which are associated with positive emotions and a sense of joy. However, the individual also struggles with school and experiences anxiety and depression, which are linked to negative emotions such as sadness, fear, and frustration.

\end{minipage}\tabularnewline \bottomrule
\end{tabular}
\end{adjustbox} \caption{Example TH-VAE timeline summary and its high-level summary. Examples for all systems in Appendix C.}
\vspace{-1em}
\label{tab:results.examples}
\end{table*}

\subsection{Evaluation}\label{evaluation} 

We use summaries by clinical experts (\S\ref{sec:experiments.dataset}) in automatic evaluation. In human evaluation we work with the same experts, where they rated summaries 
for factual consistency, salient meaning preservation, and facets of usefulness.\footnote{We merged aspects in human evaluation after a pilot, based on expert feedback. Given the LLM's ability to output well-formed text, the cognitively taxing nature of the task, and time constraints, we prioritised aspects that demand domain expertise rather than general linguistic quality (e.g. fluency).} Details on procedure and metrics are in Appendix \ref{sec:appendix.eval.auto}-\ref{sec:appendix.eval.human}. 


\smallskip
\noindent\textbf{Salient information preservation.} We adapt MHIC \cite{Srivastava20222Counseling} to assess whether timeline summaries capture clinically relevant information. Given evidence $E$ and timeline summary sentences $T$, we average the maximum recall-oriented BERTScore \cite{Zhang*2020BERTScore}:
\vspace{-.5em}
\begin{equation*}
\text{MHIC}_{sem} = \frac{1}{|E|} \sum_{e \in E} \max_{t \in T} R_{\text{BERT}}(e, t)
\end{equation*}

\noindent\textbf{Factual consistency.} 
To measure whether timeline summaries are consistent with original timelines, we apply the faithfulness score used in traditional summary evaluation with a modified procedure that splits timelines into chunks. Given a chunked timeline $D$ and its timeline summary $T$, for every sentence $t$ in $T$, we calculate the maximum probability of a timeline chunk $d$ in $D$ entailing $t$ using a NLI model and average across all summary sentences. 

\vspace{-1.5em}
\begin{equation*}
\text{FC}_\text{Timeline} = \frac{1}{|T|} \sum_{t \in T} \max_{d \in D} \text{NLI}(\text{Entail}| d, t)
\end{equation*}
\vspace{-.75em}

\noindent Next we assess the consistency of high-level model-generated summaries $S$ with human-written ones $G$, where consistency is the absence of contradiction. We define $C$ to be a function that quantifies the consistency of text $B$ based on text $A$: 

\vspace{-1.5em}
\begin{equation*} 
\resizebox{1.05\hsize}{!}{$
C(A, B) = \frac{1}{|A| \cdot |B|} \sum_{a \in A} \sum_{b \in B} \left(1 - \text{NLI}(\text{Contradict}|a, b)\right)
$}
\end{equation*}

\noindent We calculate the consistency of high-level summaries to gold summaries as $\text{FC}_\text{Expert} = C(G, S)$.

\smallskip
\noindent\textbf{Evidence appropriateness.} We measure the consistency of high-level summaries $S$ to their accompanying timeline summaries $T$ via $\text{EA} = C(T, S)$. 

\smallskip

\noindent\textbf{Coherence.} 
We estimate how easy it is to follow the summary and how effectively the mental health summary integrates information from the timeline summary using BARTScore \cite{NEURIPS2021_e4d2b6e6}
. We evaluate \emph{logical} coherence via intra-summary NLI (IntraNLI), taking the mean consistency of each sentence against all other sentences to assess the logical interconnection of information within the mental health summary. 

\smallskip

\noindent\textbf{Fluency.}  
We separately estimate fluency for timeline and high-level summaries using perplexity (PPL) under \textsc{GPT-2-xl} \cite{radford2019language}.

\smallskip

\noindent\textbf{Usefulness.} Summaries should help the clinician understand the client’s condition. This is assessed via human evaluation only, with respect to general usefulness and specific categories (diagnosis, intra- and interpersonal patterns and MoC). Details are available in the Appendix in Table \ref{tab:appendix.clinical_concepts}.

\section{Results}
\label{sec:results}

\subsection{Automatic evaluation}
Table \ref{tab:results.examples} shows example summaries. 
We perform two-tailed permutation tests in our comparisons reporting statistical significance at $\alpha=.05$. 
\begin{table}[]

\begin{adjustbox}{width=\columnwidth}
\begin{tabular}{@{}llrrrr@{}}
\toprule
Aspect & Metric & LLaMA & TH-VAE & Skeleton & Naive \\ \midrule
SMP 
    & $\text{MHIC}_\text{sem}$ & .65 & \textbf{.66} & .57 & --\\
\multirow[t]{3}{*}{FC} 
    & $\text{FC}_\text{Timeline}$ & \textbf{.63} & \textbf{.63} & .21 & --\\
    & $\text{FC}_\text{Expert}$ & .95 & \textbf{.96} & .95 & .93\\
\multirow[t]{2}{*}{EA} 
    & $\text{EA}$ & \textbf{.97} & \textbf{.97} & .95 & --\\
\multirow[t]{2}{*}{Coherence} 
    & IntraNLI & .95 & \underline{\textbf{.96}} & .95 & .93\\
    & BARTScore & \textbf{-2.96} & -3.10 & -3.09 & --\\
\multirow[t]{2}{*}{Fluency} 
 & $\text{PPL}_\text{Timeline}$ ($\downarrow$) & \underline{\textbf{13.80}} & 56.33 & 31.82 & --\\
 & $\text{PPL}_\text{High-level}$  ($\downarrow$)& 9.32 & \textbf{9.30} & 9.45 & 11.38 \\ \hline 
\end{tabular} 
\end{adjustbox}
\caption{Automatic evaluation for salient meaning preservation (SMP), factual consistency (FC), evidence appropriateness (EA), coherence, and fluency. Higher is better, except for PPL. BARTScore uses log likelihood, hence higher (less negative) is better. Best in \textbf{bold}, significant improvement over second-best \underline{underlined}. 
} 
\label{tab:results.auto}
\vspace{-1.5em}
\end{table}
TH-VAE and LLaMA generated significantly higher quality summaries compared to other baselines. TH-VAE and LLaMA were comparable on most metrics, preserving mental health information ($\text{MHIC}_{sem}$) while similarly consistent with the source ($\text{FC}_\text{Timeline}$) in timeline summaries and factually consistent with human-written references in high-level mental health summaries ($\text{FC}_\text{Expert}$). 

Two-tailed permutation tests showed that LLaMA timeline summaries were significantly more fluent ($\text{PPL}_\text{Timeline}$), in line with its tendency to normalise text (see examples, Appendix B). These tests also indicate that high-level summaries were comparably coherent in terms of ease of reading and integrating information from timeline summaries (BARTScore). This is expected since all methods used the same prompting framework to generate high-level summaries. However, TH-VAE achieved significantly higher IntraNLI, suggesting its timeline summaries allow for more logically coherent synthesis of detailed clinical information. 


\subsection{Human evaluation}
We selected three systems for human evaluation: LLaMA, TH-VAE, and the naive LLaMA baseline. This allows us to compare top-performing models and understand how removing timeline summarisation and clinical prompting steps may impact summary quality perceived by human judges.
TH-VAE produced summaries considered the most factually consistent and useful in summarising changes (MoC) among compared models. Human judges found LLaMA summaries generated with clinical prompts to be most useful in other usefulness criteria, whereas LLaMA with a simple summarisation prompt was consistently least useful. Notably, LLaMA summaries \emph{without} clinical prompts were rated as more factually consistent than those \emph{with} clinical prompts, suggesting they adhered to the source timeline, but were impacted by lack of guidance (Table \ref{tab:results.likert}). A detailed qualitative evaluation in Appendix A.6 shows that Llama timelines present more hallucinations than TH-VAE.
\begin{table}[!htbp]
\centering
\begin{adjustbox}{width=.82\columnwidth}
\begin{tabular}{@{}lccc@{}}
\toprule 
Aspect                                      & LLaMA & TH-VAE & Naive \\ \midrule
Factual Consistency                         & 3.08 & \textbf{3.35}  & 3.28    \\
Usefulness (General)      & \textbf{3.38} & 3.28  & 2.55 \\
\phantom{---} (Diagnosis) & \textbf{3.40} & 3.25 & 2.93  \\
\phantom{----}(Inter-\& Intrapersonal) & \textbf{3.48}  &  3.33 & 2.23    \\
\phantom{----}(MoC)       & 3.30   &  \textbf{3.35} & 1.18  \\ \bottomrule
\end{tabular}
\end{adjustbox}
\caption{Human evaluation results based on 5-point Likert scales (1 is worst, 5 is best). Best in \textbf{bold}.}
\label{tab:results.likert}
\vspace{-1em}
\end{table}

\vspace{-0.1cm}
\subsection{Ablation}

We performed ablation studies to investigate the importance of key phrases (\S\ref{method:key}) and elaborate clinical prompts for the final summary generation (\S\ref{method:summary}) in TH-VAE and LLaMA. Details are in Appendix \ref{ap:ablation}, Tables~\ref{tab:ablation} and \ref{tab:ablation_llama}.  We experimented with (a) removing keyphrases but keeping the clinical prompts 
and (b) keeping the keyphrases, but prompting the LLM to summarise the high-level summary directly without any guiding topics. 

In both systems, removing keyphrases results in timeline summaries capturing less salient information ($\text{MHIC}_{sem}$), and degraded logical connectedness ($\text{IntraNLI}$), evidence appropriateness ($\text{EA}$), and factual consistency with gold summaries ($\text{FC}_\text{Expert}$), showing that \emph{keyphrases help focus generation on mental health related information}. In TH-VAE, removing keyphrases made timeline summaries less consistent with the source ($\text{FC}_\text{Timeline}$), and we observed the same trend to a greater effect when clinical prompts are removed. Thus, \emph{the elaborate prompt does provide an efficient clinical guidance for the LLM to generate summaries}.

In LLaMA, removing keyphrases improves timeline summary faithfulness ($\text{FC}_\text{Timeline}$) at the expense of clinical information  ($\text{MHIC}_{sem}$). This shows the role of keyphrases guided by domain expertise as anchors in summaries of long texts. Consistency with experts ($\text{FC}_\text{Expert}$) are similar across ablation settings but highest when both are employed, underlining the importance of using these components in conjunction.
\vspace{-0.3cm}
\section{Conclusions}
\vspace{-0.3cm}
We present a novel method for hybrid abstractive summarisation using hierarchical VAE and LLMs and the first approach to creating clinically meaningful mental health summaries from users' social media timelines.
Our approach results in summaries with a dual narrative perspective: high-level third person information useful for clinicians is combined with first person corresponding evidence from users' timelines. Abstractive timeline summarisation is performed by three different systems (LLM-, TH-VAE- and skeleton-based) whose generation is guided by key-phrases obtained by an LLM through instruction prompting. High-level clinical summaries in third-person are generated by feeding the timeline summaries from all three systems into an LLM. Our proposed timeline summariser, TH-VAE, based on a hierarchical VAE for long texts, can capture long dependencies between sub-timelines and while LLM timeline summaries are the most fluent, they lag behind TH-VAE on logical coherence and factuality.
From a clinical psychology viewpoint our work enables clinician access to consented clients' social media data allowing them to understand changes in their mental state over time. Importantly it enables generation of automated summaries emphasizing essential clinical concepts which can aid mental health professionals to quickly grasp an individual's psychological condition and progression.

\section*{Limitations}
Our work considers the segmentation of timelines in terms of
moments of change as changes in an individual’s
mood judged on the basis of their self-disclosure
of their well-being. This is faced by two limiting
factors: (a) users may not be self-disclosing important aspects of their daily lives and (b) while also \cite{hills-etal-2023-creation} segment user timelines based on moments of change in mood there may be other appropriate ways to effectively segment timelines into semantically related temporal units. For example timelines could be segmented based on symptoms or life events which could also be evolving over time. Empirically we have not found topics to be an effective way of identifying sub-timelines and segments within a timeline but the best way of segmenting the timelines is an open research direction. 

Though our models could be tested in cases of nonself-disclosure (given the appropriate ground truth labels), the analysis and results presented in this
work should not be used to infer any conclusion on such cases. 

While we believe our methods for clinically meaningful longitudinal summarisation of social media data for mental health monitoring to be applicable to non-social media longitudinal data such as therapy sessions, this remains future work.

In the present study, we have conducted a comparison between timeline summarization using TH-VAE, skeleton-based and LLM-generated summaries. A further qualitative evaluation by a senior clinical therapist found that the summaries generated by Llama often reached conclusions that were not sufficiently supported by the evidence provided in the timeline, and were lower in factual consistency than the TH-VAE. The TH-VAE and Llama were effective in summarizing the intrapersonal and interpersonal patterns and moments of change, but their depiction of diagnostic aspects was only moderately accurate, characterized by some inaccuracies and omissions. These findings will help pinpoint areas where our models can be enhanced and refined. 
\section*{Ethics Statement}
Ethics institutional review board (IRB) approval
was obtained from the corresponding ethics board
of the lead University prior to engaging in
this research study. Our work involves ethical considerations around the analysis of user generated
content shared on a peer support network (TalkLife). A license was obtained to work with the user
data from TalkLife and a project proposal was submitted to them in order to embark on the project.
The current paper focuses on the summarisation of users' social media timelines for mental health monitoring, by using
moments of change (MoC) in mood as the anchors to segment timelines. These changes involve
recognising sudden shifts in mood (switches or escalations). Expert clinical annotators were 
paid fairly in line with University payscales. They
were alerted about potentially encountering disturbing content and were advised to take breaks.
The annotations are used to provide examples to an in house LLMand evaluate natural language processing models for creating mental health summaries based on users social media timelines. Working with datasets such as TalkLife
and data on online platforms where individuals
disclose personal information involves ethical considerations \cite{maoloose11, kekulluoglu2020analysing}. Such considerations include careful analysis
and data sharing policies to protect sensitive personal information. The data has been de-identified
both at the time of sharing by TalkLife but also by
the research team to make sure that no user handles
and names are visible. Any examples used in the
paper are paraphrased (generated summaries). Potential risks from the application of our work in being
able to summarise the mental health of individuals based on their social media
timelines are akin to those in earlier work on personal event identification from social media and
the detection of suicidal ideation. Potential mitigation strategies include restricting access to the code
base and corpus used for evaluation by requiring an NDA, as with other mental health datasets.

The final high level summaries in all cases are obtained by feeding the timeline summaries into an LLM. Given that LLMs are susceptible to factual inaccuracies, often referred to as 'hallucinations,' and tend to exhibit biases, the clinical summaries they generate may contain errors that could have serious consequences in the realm of mental health decision-making. These inaccuracies can encompass anything from flawed interpretations of the timeline data to incorrect diagnoses and even recommendations for potentially harmful treatments. Mental health professionals must exercise caution when relying on such generated clinical summaries. These summaries should not serve as substitutes for therapists in making clinical judgments. Instead, well-trained therapists must skillfully incorporate these summaries into their clinical thought processes and practices.
Significant efforts are required to establish the scientific validity of the clinical benefits offered by these summaries before they can be integrated into routine clinical practice.

\bibliography{anthology,custom}
\bibliographystyle{acl_natbib}

\appendix
\section{Appendix} \label{sec:appendix}

\subsection{Experimental Settings} \label{exp:setting}
\noindent\textbf{TH-VAE}
We load pre-trained parameters from \textsc{bart-base} \cite{lewis2020-bart} for pre-trained word embedding and 6 transformer decoder layers in the model. We set the dimensional size of $\boldsymbol{z}_{i}$ to be the same as the size of word embeddings (768). We set the number of latent variables $l$ as 5, which has the best performance on our dataset. In addition, we set the number of cells in \textbf{block} is 3, and the number of cells in each \textbf{group} is 1. We use the Adam optimizer \cite{kingma2014adam} (learning rate: $5$$\times$$10^{-4}$).

\smallskip
\noindent\textbf{LLM}
Our experiments use 4bit-quantized \textsc{llama-2} \citep{touvron2023llama}. For keyphrase extraction, we use few-shot prompting on the base pre-trained model \textsc{llama-2-13b}. In zero-shot prompting tasks with detailed instructions (i.e. mental health related inferences), we use the chat version of the model \textsc{llama-2-13b-chat} to take advantage of its fine-tuning on instruction datasets and human preferences. 

We trained TH-VAE with 2 hours on 1 GPU, and spent 20 GPU hours for generating high-level summaries.

\subsection{Evaluation Metrics} \label{sec:appendix.eval.auto}
\paragraph{NLI} On metrics that require NLI, we use a \textsc{RoBERTa} model \citep{liu2020roberta} fine-tuned on fact verification and NLI\cite{nie-etal-2020-adversarial}: \url{https://huggingface.co/ynie/roberta-large-snli_mnli_fever_anli_R1_R2_R3-nli}. When evaluating evidence appropriateness, we consider text from the timeline summary to be the premise and text from the high-level summary to be the hypothesis. When running the NLI model, we prefix every sentence in the timeline summary with "The individual wrote:". While we did not find statistically significant differences between the selected prefix vs. no prefix and vs. similar alternatives, we decided on prefixing as empirically it seemed to help the NLI model on noisy premises.

\paragraph{Salient Meaning Preservation: MHIC}
We make the following changes to MHIC \cite{Srivastava20222Counseling}. First, instead of ROUGE we measure semantic embedding similarity using BERTScore. Second, instead of computing separate scores based on hard utterance categories, we compute a unified one using the semantic intersection of information highlighted by annotators. 

We find the intersection of highlighted timeline spans among annotators by (1) directly extracting intersecting substrings, (2) computing pairwise cosine similarity across evidence spans, keeping pairs with similarity >= .60, then selecting the shorter span from each pair, and (3) deduplicating evidences from these steps. We use the sentence-transformers library and \textsc{msmarco-distilbert-base-v3} embeddings.

\paragraph{Factual Consistency} For $\text{FC}_\text{Timeline}$, we chunk timeline texts with a cutoff of 60 tokens to match input lengths in the NLI model's training data. 

\subsection{Annotation \& Human Evaluation} \label{sec:appendix.eval.human}
\paragraph{Training}
We ran training sessions for both summarisation and evaluation tasks under the supervision of a senior clinical expert to ensure annotators clearly understood task requirements. 

\paragraph{Summarisation}
During the training session, the annotation team were introduced to the dataset and task, and were provided with guidelines. After reviewing the guidelines and held out examples, we worked on a timeline reserved for annotator training together. The annotators separately worked on another timeline reserved for training. We compared annotations during the second training session. Once we were confident that the team had a shared understanding of the task requirements, the annotators proceeded to actual timelines used for testing in this paper.

\paragraph{Evaluation}
We provided the annotators with guidelines and introduced the evaluation task as well as criteria (see Appendix D) in the first training session. We checked agreement on a small set of timelines, then after discussion and clarifications on a second session they were asked to proceed to rating summaries on the remaining test timelines. 

During evaluation, annotators were presented data on a timeline-by-timeline basis. When rating summaries for a timeline, they would receive the summaries in a randomly shuffled order. 

\subsection{Ablation Results}\label{ap:ablation}
\begin{table}[!htbp]
\begin{adjustbox}{width=\columnwidth}
\begin{tabular}{@{}llrrr@{}}
\toprule
Aspect & Metric & TH-VAE & \makecell[c]{\textit{-keyphrases}} & \makecell[c]{ \textit{-clinical prompts}}  \\ \midrule
SMP 
    & $\text{MHIC}_\text{sem}$ & \textbf{.66} & .62 & -- \\
\multirow[t]{3}{*}{FC} 
    & $\text{FC}_\text{Timeline}$ & \textbf{.63} & .52 & -- \\
    & $\text{FC}_\text{Expert}$ & \textbf{.96} & .95 & .91 \\
\multirow[t]{2}{*}{EA} 
    & $\text{EA}$ & \textbf{.97} & .94 & .93 \\
\multirow[t]{2}{*}{Coherence} 
    & IntraNLI & \textbf{.96}& .95 & .94 \\
    & BARTScore & -3.10 & -3.08 & \textbf{-2.74} \\
\multirow[t]{2}{*}{Fluency} 
 & $\text{PPL}_\text{Timeline}$ ($\downarrow$) & \textbf{56.33} & 81.45 & -- \\
 & $\text{PPL}_\text{High-level}$  ($\downarrow$)& \textbf{9.30} & 9.38 & 13.62 \\ \hline 
\end{tabular} 
\end{adjustbox}
\caption{Ablation results. Best in \textbf{bold}. TH-VAE without clinical prompts uses the same timeline summary as TH-VAE so repeated metrics were removed for brevity. } 
\label{tab:ablation}
\end{table}

\begin{table}[] 
\begin{adjustbox}{width=\columnwidth}
\begin{tabular}{@{}llrrrr@{}}
\toprule
Aspect & Metric & LLaMA & \makecell[c]{\textit{-keyphrases}} & \makecell[c]{\textit{-clinical prompts}} & Naive  \\ \midrule
SMP 
    & $\text{MHIC}_\text{sem}$ & \textbf{.65} & .59 & -- & -- \\
\multirow[t]{3}{*}{FC} 
    & $\text{FC}_\text{Timeline}$ & .63 & \textbf{.68} &  -- & --\\
    & $\text{FC}_\text{Expert}$ & \textbf{.95} & .93 & .93 & .93\\
\multirow[t]{2}{*}{EA} 
    & $\text{EA}$ & \textbf{.97} & .93 & .94 & -- \\
\multirow[t]{2}{*}{Coherence} 
    & IntraNLI &  \textbf{.95} & .89 & .90 & .93 \\
    & BARTScore & -2.96 & \textbf{-2.48} & -2.61 & -- \\
\multirow[t]{2}{*}{Fluency} 
 & $\text{PPL}_\text{Timeline}$ ($\downarrow$) & 13.80 & \textbf{11.38} & -- & -- \\
 & $\text{PPL}_\text{High-level}$  ($\downarrow$)& \textbf{9.32} & 13.78 & 11.62 & 11.38 \\ \hline 
\end{tabular} 
\end{adjustbox}
\caption{Ablation results. Best in \textbf{bold}. LLaMA without clinical prompts uses the same timeline summary as LLaMA so repeated metrics were removed for brevity. Naive uses neither keyphrases nor clinical prompts.}

\label{tab:ablation_llama}
\end{table}

\subsection{Clinical Concepts}

\begin{table}[H]
\centering
\begin{adjustbox}{width=\columnwidth}

\begin{tabular}{l}
\hline
\textbf{Diagnosis} \\
\hline
Presenting issues \small{(what bothers the person and causes distress; triggers)}. \\
Mental health symptoms, level of functioning, well-being. \\
Physical symptoms. \\
Risk assessment \small{(previous suicidal attempts, intent to suicide, access to} \\ \small{lethal means; hopelessness, social isolation, recent loss, impulsivity,} \\ \small{dramatic mood swings)}. \\
Motivation to change.  \\
Lifestyle \small{(diet, physical activity, sleep, alcohol/drug/tobacco use,} \\ \small{occupation, environment, screen time, healthcare practices)}. \\
Agency, coping mechanisms, strengths and resources \small{(what helps } \\ \small{the person, how they typically cope with stress and difficulties, resilience)}. \\
Meaning/goals/direction in life. \\
Behaviour \small{(adaptive and maladaptive behavioural patterns)}. \\
Important events \small{(present and past events in life; traumatic events)}. \\
\hline
\textbf{Intrapersonal and Interpersonal patterns} \\
\hline
Main need/wish/desire. \\
Interpersonal relationships \small{(repetitive interpersonal pattern;} \\ \small{conflicts; how others are perceived; social support)}. \\
Self perception, self esteem. \\
\hline
\textbf{Moments of change} \\
\hline
Emotion \small{(sad, happy, etc)}. \\
Arousal level \small{(high/low)}. \\
Emotion regulation strategies. \\
Switches \small{(drastic change of one's mood)}. \\
Escalations \small{(intensification in one's mood)}. \\
Self understanding \small{(insights about the self and the relationship; ability to}\\ \small{reflect and understand repetitive patterns)}. \\
\hline
\end{tabular}
\end{adjustbox}
\caption{Clinical concepts important to therapeutic approaches. Our task is to capture and summarize them if such information is present in user timelines.}
\label{tab:appendix.clinical_concepts}
\end{table}

\subsection{Qualitative discussion of clinical summaries} \label{sec.appendix.qualitative}
The TH-VAE and Llama-best models offered moderately insightful details regarding the individual's diagnosis. Their summaries accurately captured the general aspects of the diagnosis, focusing mainly on evident symptoms while overlooking some critical elements. The Llama-best model often reached conclusions that were not sufficiently supported by the evidence provided in the timeline. For example, both models noted the individual's depression, self-harm, and suicidal thoughts but failed to recognize a clear eating disorder. Additionally, the Llama-best suggested PTSD without substantial evidence in the provided timeline. However, these models were useful in shedding light on the individual's self and relational dynamics over time. In contrast, the basic-prompt model presented a very broad summary, missing several vital details and failing to reflect significant clinical concepts. On the other hand, the TH-VAE and Llama-best produced more comprehensive summaries, effectively highlighting crucial aspects of the individual's self-perception, interpersonal relationships, and moments of change. Overall, from a clinical point of view, the quality of the summaries generated by the TH-VAE and Llama-best models were quite similar. The Llama best was only slightly lower in factual consistency than the TH-VAE. The TH-VAE and Llama-best models were effective in summarizing the intrapersonal and interpersonal patterns and moments of change, but their depiction of diagnostic aspects was only moderately accurate, characterized by some inaccuracies and omissions.
\clearpage


\section*{Supplementary material}

# Appendix B. Instruction Prompts
## B.1 Keyphrase Extraction

```
Task: Choose key phrases in the following posts.
Text: {example post 1}
Keyphrases:{expert key phrases 1}
Text: {example post 2}
Keyphrases:{expert key phrases 2}
Text: {concatenated posts to annotate}
Keyphrases:
```

## B.2 Timeline Summarisation: LLaMA

```
Write a TLDR as the user (first-person), focusing on the keyphrases.
Keyphrases: {extracted keyphrases}
{concatenated posts to summarise}
TLDR:
```

## B.3 High-level only: LLaMA Naive Baseline

```
You are a helpful assistant to an expert therapist who reads social media
chronological text written by an individual who has mental health concerns.

Summarize the texts below:
{timeline chunk concatenated}
```

## B.4 Map Prompt: Diagnosis

```
Your goal is to describe the individual's mental state and identify potential
indicators that may suggest a mental health diagnosis, considering the following
aspects:
1. Presenting Issues: What are the main concerns or stressors evident in the
individual's posts? What triggers seem to affect their mental state?
2. Mental Health Symptoms and Functioning: Does the individual exhibit any mental
health symptoms? How are their mood, energy levels, and interest in usual
activities? Are there noticeable changes in sleep patterns, appetite,
concentration, or social interactions? How do they describe their overall well-
being and functioning in daily activities?
3. Mental Health Treatment History: Has the individual been in contact with
mental health professionals such as psychiatrists or psychotherapists? Are there
mentions of current or past outpatient or inpatient mental health treatments? Do
they reference taking psychiatric medications?
4. Physical Health: Are there any current or past physical health issues, medical
conditions, hospitalizations, or surgeries mentioned?
5. Risk Assessment: Is there evidence of previous suicidal attempts or current
suicidal thoughts? Do they have access to lethal means? What level of
hopelessness is expressed? Do they discuss social isolation, recent losses,
impulsivity, or dramatic mood swings?
6. Lifestyle Factors: What do the individual's posts reveal about their lifestyle
habits, such as diet, physical activity, sleep patterns, occupation, environment,
screen time, and healthcare practices?
7. Substance Use: Are there any references by the individual to the use of
substances like alcohol, drugs, or tobacco? If so, how frequently do they use
these substances?
8. Significant Life Events and Family History: Are there references to
significant life events like divorce, loss of a close person, experiences of
abuse, or neglect? Is there any mention of psychiatric problems or treatments
among family members?
9. Motivation and Coping Strategies: What does the individual express about their
motivation for change? How do they cope with stress and difficulties? What
strengths and resources do they have? What seems to help them? How resilient do
they appear? Do they discuss having direction, meaning, or goals in their life?

You must not make anything up. Keep the description concise and only describe
observations if they are fully supported by the text.

Here are the texts:
{Timeline summary}
```

## B.5 Map Prompt: Intrapersonal and Interpersonal Patterns

Your goal is to identify the person's main intrapersonal and interpersonal pattern, considering the following aspects:
1. Wish/Need/desire/intention/expectation:  What is the person's most dominant need, desire, intention, expectation from others and from themselves? Are there any other needs or wishes that might be indicated in a less obvious way? How well does the individual communicate their needs/ wishes with others?
2. Response of Others: How does this person typically perceive the emotions, behaviors, and thoughts of others? Are there any other perceptions of the other that might be indicated in a less obvious way? Is the individual capable of acknowledging the complex nature of others?
3. Response of self to others: How does this person tend to feel and react to others? Are there any other reactions towards others that might be indicated in a less obvious way?
4. Response of self to self: What is the individual's most dominant emotion, behavior and cognition toward oneself? Are there any other emotions and cognitions towards the self that might be indicated in a less obvious way? What is the level of self-compassion and acceptance of strengths and vulnerabilities?
5. Patterns: What is this individual's predominant dysfunctional intrapersonal and interpersonal pattern? What is this individual's predominant adaptive intrapersonal and interpersonal pattern?

You must not make anything up. Keep the description concise and only describe observations if they are fully supported by the text.

Here are the texts:

{Timeline summary}

## B.6 Map Prompt: Moments of Change (MoC)

Your goal is to understand and summarise changes over time in this individual's mood, well-being and functioning (individual/self well being; interpersonal well-being including family and close relationships; social well-being, including work, school and friends). From the following text, identify whether there are changes in the individual's expressed mood, well-being, and functioning.
a. Overall, does the mood/ well being/ functioning stay consistent or fluctuate over time?
b. Are there specific positive to negative changes? Which events (personal or interpersonal) are associated with these changes?
c. Are there specific negative to positive changes? Which events (personal or interpersonal) are associated with these changes?

Answer each concisely and reply None if not available. You must not make anything up. Keep the description concise and only describe observations if they are fully supported by the text.

Here are the texts:

{Timeline summary}

## B.7 Reduce Prompt

Rewrite these sentences about the individual so that they form a single clear and coherent document. Combine redundant sentences, but do not remove any clinical information or information about interpersonal relationships. You must not make anything up.

{Previously generated summaries}

## Appendix C. Example Summaries

### C.1 Timeline Summaries

| | |
|---|---|
| TH-VAE | I hate school. im so scared.i love the cat so much. i'm gonna go be counter for a while. unknown good night all! i hate my nose. fuck the sat. i got a commitment ring 3 i've got counterbodied under my eyes. i feel alone i feel like everyone hates me. i hate saying this because it sounds bad, but i really want someone to pay attention to me. solid my mom only cares about my brothers learning disabilities and her new boyfriend expressive. my boyfriend goes throug of ptsd like depression symptoms, so he gets distant sometimes levels. i literally have no friends argue. i've always been quiet but sometimes i just want someone never pay attention to me. i argue to know only when't be a body. i have to know. i'll have to be happy. |
| LLaMA | Need help passing a breathalyzer test for homecoming, worried about proposal, struggling with school, and feeling insecure about appearance. I love my cat so much, but I have no motivation and can't get out of bed. I really enjoy making others happy, but I struggle with feeling alone and like everyone hates me. What is a healthy weight for a 5'2\" 17 year old girl? Is there anything besides vodka that can help me get through homecoming? Someone please pay attention to me. I feel ignored by my family and my chat isn't working. I have no friends and my boyfriend has PTSD-like symptoms. I promise I'm not ignoring you, I just need someone to talk to. My cat snoozes and it's wonderful. "I just wanna be kissed" but I hate my nose and have dark circles under my eyes. |
| Skeleton | don't know what to do anymore i don't want to go to homecoming but i can't help my own time i feel like i'm going to try to share my feelings but i know that i'm feeling alone but i just want someone to  pay attention to me. i love my cat so much. i can tell you that i feel good at the same time i'm in love with my partner and i'm glad i could turn up to be my best friend and i know what it happened to me i bet you're going to make me feel better cause i'm sure if i can turn my homecoming if i do help i will be happy but i like a lot of people i want to tell you i appreciate you but i won't help you just want you to go to sleep but i didn't really want to talk to you my way to help my mom so i can help you and make you feel better at the end of my life. i love him so much and i bet it's hard to me but i hope you were a good time but i need to be able to see you but you don't like you but it's okay to see your |

## C.2 High-level Summaries

| | |
|---|---|
| TH-VAE | The individual is likely experiencing a range of mental health issues, including depression, anxiety, and low self-esteem. They report feeling isolated, lonely, and frustrated, with a history of stressful life events. The individual's mood appears to be low, with expressions of sadness, frustration, and anger.<br><br>It is evident that they have a profound craving for attention and connection with others, as well as a yearning for acceptance and validation. However, their fear of rejection and abandonment hinders them from forming and maintaining healthy relationships. Moreover, their self-criticism and preoccupation with perceived shortcomings indicate a lack of self-compassion and acceptance of their strengths and vulnerabilities.<br><br>The individual's mood and well-being have been observed to fluctuate over time, with both positive and negative changes experienced. The individual has expressed a range of emotions, including sadness, loneliness, and frustration, as well as moments of happiness and positivity. Noteworthy positive changes include their excitement about having a cat and receiving a commitment ring, which are associated with positive emotions and a sense of joy. However, the individual also struggles with school and experiences anxiety and depression, which are linked to negative emotions such as sadness, fear, and frustration. |
| LLaMA | Based on the information provided, the individual is experiencing a range of mental health symptoms, including depressive and anxiety symptoms, low self-esteem, and suicidal ideation, which are likely interfering with their daily functioning and quality of life. Addressing lifestyle habits such as excessive screen time and lack of physical activity may also be helpful in managing symptoms.<br><br>The individual's primary intrapersonal and interpersonal patterns appear to be a mix of dependent and self-centered tendencies, with a lack of self-awareness and self-regulation. These patterns are characterized by negative self-talk, low self-esteem, and a fear of being alone, leading to a reliance on alcohol to cope with stress. The individual also exhibits difficulty regulating their emotions and behaviors, and seeks validation and attention from others while having difficulty setting healthy boundaries.<br><br>The individual experiences fluctuating mood and well-being, with feelings of worry, insecurity, and loneliness, as well as moments of joy and happiness when they make others happy. Despite these positive interactions, the individual struggles with negative thoughts and feelings of being ignored by family and friends. They desire positive changes and may benefit from seeking professional help to address symptoms of depression, anxiety, and low self-esteem. |

| | |
|---|---|
| Skeleton | Based on the provided social media texts, the individual is struggling with multiple mental health issues, including depression, anxiety, low self-esteem, and possible borderline personality disorder. Their mood appears low, with expressions of feeling alone, lost, and tired, which may indicate depression and anxiety. The individual's desire to "feel better" and "turn up" suggests a willingness to seek help and improve their mental state. The lack of direct references to mental health treatment history and their expressed desire for improvement suggest that they may be open to seeking professional help. Their statements about feeling alone and lost, as well as their desire to "turn up," may indicate a higher risk of suicide.<br><br>The individual longs for human connection and yearns for validation from others, yet they feel overwhelmed by their own emotions. They find solace in their attachment to their cat and appreciate their partner's support. However, they still experience a sense of loneliness and disconnection, which they introspectively examine. The individual's primary intrapersonal patterns may be characterized by isolation and disconnection, while their adaptive patterns may involve self-awareness and introspection.<br><br>The individual's mood and well-being appear to fluctuate over time, with both positive and negative emotions being experienced. They express feelings of happiness and appreciation for their partner and pets, but also mention feeling alone and isolated, which may suggest a decline in their well-being. The text is written in a stream of consciousness style, providing a glimpse into the individual's inner thoughts and emotions, but it does not provide a complete picture of their mood and well-being. |
| LLaMA Naive | Based on the information provided, it appears that the individual is seeking emotional support and connection. They mention feeling alone and isolated, and express a desire for someone to pay attention to them and listen to their needs. The individual is feeling overwhelmed and frustrated, expressing a desire to leave and pursue their dreams. They are struggling with negative self-talk and body image issues, feeling ignored and alone despite expressing happiness and positivity. They use humor and sarcasm to cope with their emotions, but also mention feeling scared and insecure. The individual is seeking alternative ways to cope with their feelings besides drinking vodka. |

## Appendix D. Human Evaluation Criteria

| | |
|---|---|
| **Factual consistency**<br><br>A factually consistent summary accurately reflects the content of the timeline. It does not contain information that is not present in the timeline. | 1 - Not at all factually consistent: The summary contains significant inaccuracies or misrepresentations, completely misaligning with the timeline's content.<br>2 - Mostly not factually consistent: The summary contains significant inaccuracies or misrepresentations, poorly reflecting the timeline's content.<br>3 - Somewhat factually consistent: The summary is somewhat accurate, with several inaccuracies or omissions, but retains a basic reflection of the timeline's content.<br>4 - Mostly factually consistent: The summary is largely accurate, with minor inaccuracies or omissions that do not majorly distort overall understanding.<br>5 - Fully factually consistent: The summary is completely accurate, perfectly aligning with the timeline's content without discrepancies. |
| **General usefulness and Salient meaning preservation**<br><br>A useful summary should help the clinician understand the client's condition. It should contain the most clinically important information of the timeline. It does not include parts of the timeline that are less important. | 1 - Not at all useful: The summary fails to capture any essential information, significantly misrepresenting or omitting critical aspects of the individual's condition.<br>2 - Slightly useful: The summary includes some important details but primarily focuses on irrelevant or less critical information.<br>3 - Moderately useful: The summary captures important information but still includes less relevant details or omits minor key elements.<br>4 - Very useful: The summary highlights most of the crucial information, with only minor irrelevant details.<br>5 - Extremely useful: The summary encapsulates all critical information, providing a comprehensive and clear understanding of the individual's condition, without providing irrelevant information. |
| **Usefulness (diagnosis)**<br><br>The summary provides useful information about the individual's diagnosis (such as presenting issues, mental health & physical symptoms, risk assessment, behaviour). | 1 - Not at all useful: The summary fails to provide information regarding the individual's diagnosis, or it clearly distorts the individual's diagnosis by incorrectly identifying diagnostic elements.<br>2 - Slightly useful: The summary provides minimal information related to the individual's diagnosis. While the summary includes some correct diagnostic elements, it generally contains irrelevant or incorrect details and omissions.<br>3 - Moderately useful: The summary is generally accurate about the individual's diagnosis but only describes the more obvious aspects, with some information possibly missing or unclear.<br>4 - Very useful: The summary accurately identifies the individual's diagnosis and captures almost all the essential diagnostic information with only minor gaps.<br>5 - Extremely useful: The summary is comprehensive and accurately details all aspects of the individual's diagnosis. |

| | |
|---|---|
| **Usefulness (interpersonal and intrapersonal pattern)**<br><br>The summary provides helpful information about the individuals' main needs and patterns of self and other relationships. | 1 - Not at all useful: The summary provides no insight into the individual's interpersonal and intrapersonal patterns.<br>2 - Slightly useful: The summary provides a minimal understanding of interpersonal and intrapersonal patterns.<br>3 - Moderately useful: The summary covers some key aspects of the individual's interpersonal and intrapersonal patterns but may lack depth or miss important elements.<br>4 - Very useful: The summary provides a comprehensive overview of the individual's interpersonal and intrapersonal patterns, with only slight gaps or generalizations.<br>5 - Extremely useful: The summary gives a detailed and complete understanding of the individual's interpersonal and intrapersonal patterns. |
| **Usefulness (moments of change)**<br><br>The summary provides useful information about the individual's changes over time in emotion/cognition and behaviour. Where appropriate, it should help connect information between events and the individual's responses. | 1 - Not at all useful: The summary fails to provide any accurate information about whether there are changes in the individual over time.<br>2 - Slightly useful: The summary includes information about changes, but they are generally inaccurate and overlook key developments/connections, or they generally contain irrelevant information.<br>3 - Moderately useful: The summary accurately describes whether there are changes, although there may be some weaknesses or omissions as well as irrelevant information.<br>4 - Very useful: The summary accurately describes whether there are changes and where available offers helpful insights.<br>5 - Extremely useful: The summary accurately describes whether there are changes and where available provides clear, well-connected insights about the individual's development over time. |


\end{document}